\documentclass[sigconf]{acmart}
\usepackage{soul}
\usepackage{url}
\usepackage{color}
\usepackage{graphicx}
\usepackage{amsmath}
\usepackage{amsthm}
\usepackage{algorithm}
\usepackage{algorithmic}
 
\usepackage{amsfonts,amssymb}
\usepackage{multirow}
\usepackage{subfig}
\usepackage{balance}
\usepackage{enumitem}
\AtBeginDocument{%
  }

\copyrightyear{2025}
\acmYear{2025}
\setcopyright{acmlicensed}\acmConference[WWW '25]{Proceedings of the ACM Web Conference 2025}{April 28-May 2, 2025}{Sydney, NSW, Australia}
\acmBooktitle{Proceedings of the ACM Web Conference 2025 (WWW '25), April 28-May 2, 2025, Sydney, NSW, Australia}
\acmDOI{10.1145/3696410.3714906}
\acmISBN{979-8-4007-1274-6/25/04}

\begin{document}

\title{Graph with Sequence: Broad-Range \\ Semantic Modeling for Fake News Detection}

\author{Junwei Yin}
\affiliation{%
  \institution{Chongqing University}
  \city{Chongqing}
  \country{China}
}
\email{junweiyin@cqu.edu.cn}

\author{Min Gao}
\authornote{Corresponding author. Code is available at https://github.com/yyy-jw/BREAK.}
\affiliation{%
  \institution{Chongqing University}
      % \institution{Laboratory of Dependable Service Computing in Cyber Physical Society (Chongqing University) Ministry of Education}
  \city{Chongqing}
  \country{China}}
\email{gaomin@cqu.edu.cn}

\author{Kai Shu}
\affiliation{%
  \institution{Emory University}
  \city{Atlanta}
  \country{USA}
}
 \email{kai.shu@emory.edu}

\author{Wentao Li}
\affiliation{%
 \institution{University of Leicester}
 \city{Leicester}
 \country{United Kingdom}}
 % \email{livent@126.com}
 \email{wl226@leicester.ac.uk}

\author{Yinqiu Huang}
\affiliation{%
  \institution{Meituan}
  \city{Chengdu}
  \country{China}}
  \email{yinqiu@cqu.edu.cn}

\author{Zongwei Wang}
\affiliation{%
  \institution{Chongqing University}
  \city{Chongqing}
  \country{China}}
\email{zongwei@cqu.edu.cn}
\renewcommand{\shortauthors}{Junwei Yin et al.}

\begin{abstract}
The rapid proliferation of fake news on social media threatens social stability, creating an urgent demand for more effective detection methods. While many promising approaches have emerged, most rely on content analysis with limited semantic depth, leading to suboptimal comprehension of news content. 
To address this limitation, capturing broader-range semantics is essential yet challenging, as it introduces two primary types of noise: fully connecting sentences in news graphs often adds unnecessary structural noise, while highly similar but authenticity-irrelevant sentences introduce feature noise, complicating the detection process.
To tackle these issues, we propose BREAK, a \underline{b}road-\underline{r}ange s\underline{e}mantics model for f\underline{ak}e news detection that leverages a fully connected graph to capture comprehensive semantics while employing dual denoising modules to minimize both structural and feature noise.
The semantic structure denoising module balances the graph’s connectivity by iteratively refining it between two bounds: a sequence-based structure as a lower bound and a fully connected graph as the upper bound. This refinement uncovers label-relevant semantic interrelations structures.
 Meanwhile, the semantic feature denoising module reduces noise from similar semantics by diversifying representations, aligning distinct outputs from the denoised graph and sequence encoders using KL-divergence to achieve feature diversification in high-dimensional space.
The two modules are jointly optimized in a bi-level framework, enhancing the integration of denoised semantics into a comprehensive representation for detection. Extensive experiments across four datasets prove that BREAK significantly outperforms existing fake news detection methods. 
% Code is available at https://github.com/yyy-jw/BREAK/.
% \href{https://anonymous.4open.science/r/BREAK}
\end{abstract}

\begin{CCSXML}
<ccs2012>
   <concept>
       <concept_id>10010147.10010178.10010179</concept_id>
       <concept_desc>Computing methodologies~Natural language processing</concept_desc>
       <concept_significance>500</concept_significance>
       </concept>
 </ccs2012>
\end{CCSXML}

\ccsdesc[500]{Computing methodologies~Natural language processing}

\keywords{Fake News Detection, Broad-Range Semantics, Bi-Level Optimization, Graph Neural Network}

\maketitle

\begin{figure}[htbp]
    \centering
    \includegraphics[width=\linewidth,height=0.22\textwidth]{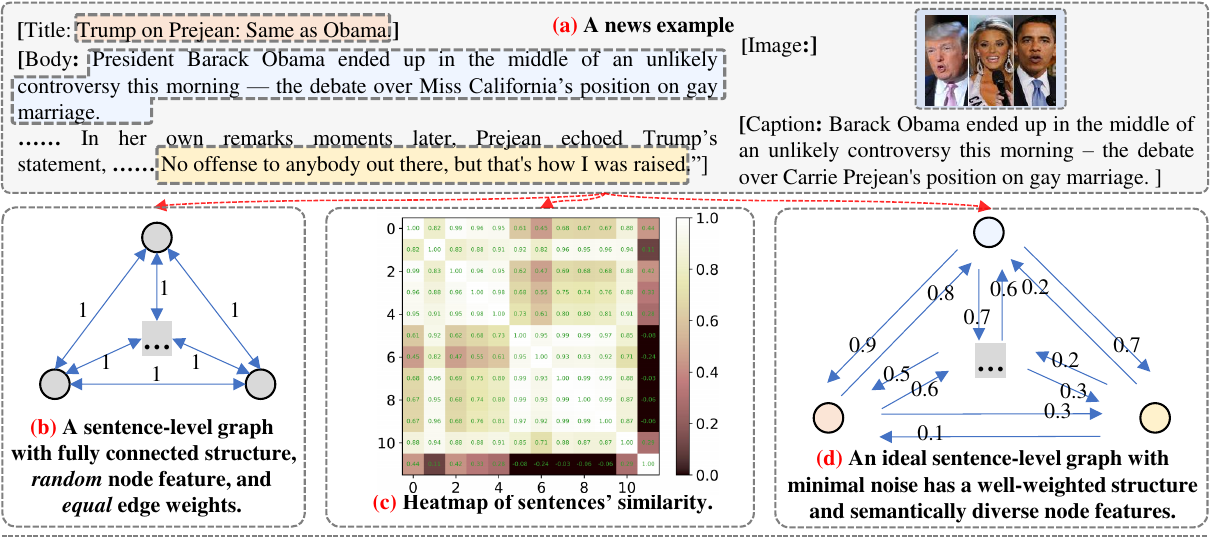}
    \caption{Comparison of traditional and denoised sentence-level graph representations. (a) A news example with both textual and visual information. (b) A traditional sentence-level graph that introduces structural noise (irrelevant connections) and overlooks the semantics of individual sentences. (c) A heatmap showing excessive similarity hindering key sentence distinction. (d) A denoised sentence-level graph (our approach), minimizing irrelevant connections and enhancing node diversity for better key sentence identification.
    }
    \label{fig1}
    % \vspace{-10pt}
\end{figure}

\section{Introduction}
\label{section1}
The Internet's rapid growth has elevated social media platforms like X (formerly known as Twitter) to vital sources of daily information. However, social media provides users freedom to create and disseminate information, making it an ``ideal'' environment for fake news dissemination \cite{Metaprompt}. The proliferation of fake news poses great threats to social safety \cite{2017social} and public health  \cite{IPM2025health}\cite{royal}, thereby necessitating fake news detection as a prominent and urgent research topic.

Throughout the development of fake news detection, news content, as the core component of news, has consistently played a pivotal role. Therefore, fully modeling news content is essential and offers distinct advantages, particularly for the early detection when only news content is available \cite{ALGM}. Moreover, it is beneficial for methods incorporating external data (e.g., comments \cite{defend}), as it provides a more comprehensive representation of news content, facilitating better integration with external data \cite{22WWWrumor}.
This has been supported by extensive research, including the use of sequence-based and graph-based models \cite{seqgraph}.
Sequence-based approaches focus on extracting sentimental, semantic, and sometimes cross-modal features from news content, particularly within news texts \cite{2021DualSenti} \cite{2023Find}\cite{IPMprogress}. 
These texts typically encompass several hundred words, often reiterating key facts and establishing a logical structure replete with clear contextual connections \cite{23EMNLP}. However, sequence-based methods, better suited for shorter texts, struggle to capture these extended, broader-range connections, overlooking essential contextual nuances \cite{22Evidence}.

In contrast, graph-based methods strive to encapsulate broad-range semantics by depicting news text as graphs at either the word level \cite{www2024msynfd} or sentence level \cite{2023TCSS}. Typically, these methods link nodes via sliding windows or fully connected structures for in-depth contextual analysis. However, they often fail to thoroughly unearth the intricate semantics embedded within complex news articles. This shortfall is particularly evident in their limited capacity to fully capture broad-range semantics and the semantic interrelations between nodes.
For instance, word-level incorporate lower noise but it also overlooks semantic interrelations outside the sliding window \cite{23EMNLP}\cite{22WWWrumor}, and generates an overwhelming number of nodes for lengthy news articles, resulting in substantial computational complexity.
A few limited studies have explored sentence-level fully connected graphs with \textit{random initialization} node features to address these shortcomings \cite{vaibhav2019}. In contrast to the word-level graph, the sentence-level graph only contains nearly one-tenth of the number of nodes and has better node semantic features.

Although studies based on fully connected graphs successfully capture broader-range semantics, they encounter a significant obstacle: while a fully connected graph structure provides comprehensive coverage, it tends to introduce noise in both the graph structure and the feature representations, such as irrelevant connections and highly similar node features, as illustrated in Figures \ref{fig1}(b) and \ref{fig1}(c). This raises the question: \textit{how can we effectively model broader range semantics while minimizing the introduction of noise?} Handling this noise is essential but comes with two critical challenges, since news texts often reiterate and emphasize key facts \cite{23EMNLP}, resulting in more complex semantic interactions and higher sentence similarity. Firstly, the semantic interrelations within a news article can be extremely complex. For example, news articles with 10 sentences could have 90 ($2\times\Sigma^{9}_{k=1}k$) directed edges, each with different weights, making it hard to eliminate noise and capture key interrelations.
Secondly, the semantics of news sentences often exhibit high similarity, hindering the differentiation of key sentences, as depicted in Figure \ref{fig1}(c). These challenges of denoising significantly impede the comprehensive extraction of news content and undermine detection performance, highlighting the need for innovative solutions.

To address the challenges of fake news detection, we introduce BREAK—a model designed to denoise and integrate \underline{b}road-\underline{r}ange s\underline{e}mantics for f\underline{ak}e news detection. BREAK effectively tackles the noise in both structural and feature semantics, which often obstruct accurate detection, by harmonizing graph-based and sequence-based representations. 
BREAK addresses structural noise through an edge weight inference mechanism in its semantic structure denoising module. It first models news content as a fully connected bidirectional graph to capture all potential semantic interrelations. Then, it refines this structure by integrating sequential semantics as a lower bound, treating the fully connected graph as an upper bound and progressively narrowing the graph based on semantically relevant connections. This dynamic process strengthens important links while filtering out irrelevant ones, thereby reducing structural noise.
To handle feature noise, BREAK employs a semantic feature diversification mechanism in the semantic feature denoising module. A graph encoder captures broad-range semantics across the entire news article, while a sequence encoder preserves more accurate sequential semantics. By aligning these two distinct representations using KL-divergence, BREAK ensures feature diversification and mitigates the noise from redundant or irrelevant sentence-level semantics.
Through this combined denoising process, BREAK produces a comprehensive and refined semantic representation, significantly enhancing its effectiveness in detecting fake news. The innovation of BREAK lies in its dual denoising approach, which balances capturing broad-range semantics with maintaining sentence-order precision, ultimately leading to more accurate detection.
Our contributions in this research are threefold:
\begin{itemize}[leftmargin=*, topsep=0pt, partopsep=0pt]
\item Broad-Range Semantic Modeling for Fake News Detection. We propose BREAK, a novel model using a fully connected graph to capture comprehensive semantics, with dual denoising modules reducing structural and feature noise for more refined, accurate semantic representation.

\item Dual Denoising Mechanisms. We design two denoising modules: the semantic structure denoising module, which iteratively refines graph connectivity between a sequence-based lower bound and a fully connected upper bound to uncover label-relevant semantics, and the semantic feature denoising module, which reduces noise from similar semantics by aligning outputs from graph and sequence encoders using KL-divergence.

\item Extensive Experimental Validation. We perform extensive experiments on four distinct datasets, showing that BREAK significantly outperforms existing methods in fake news detection. The results show the effectiveness of the bi-level framework in integrating denoised semantics and improving detection performance.
%Extensive experiments on four datasets demonstrate the superiority of BREAK. Meanwhile, we explore the effectiveness of a combination of GNN and sequential models and denoising based solely on news content.
\end{itemize}

\begin{figure*}[htbp]
    \centering
    \includegraphics[width=\linewidth,height=0.31\textwidth]{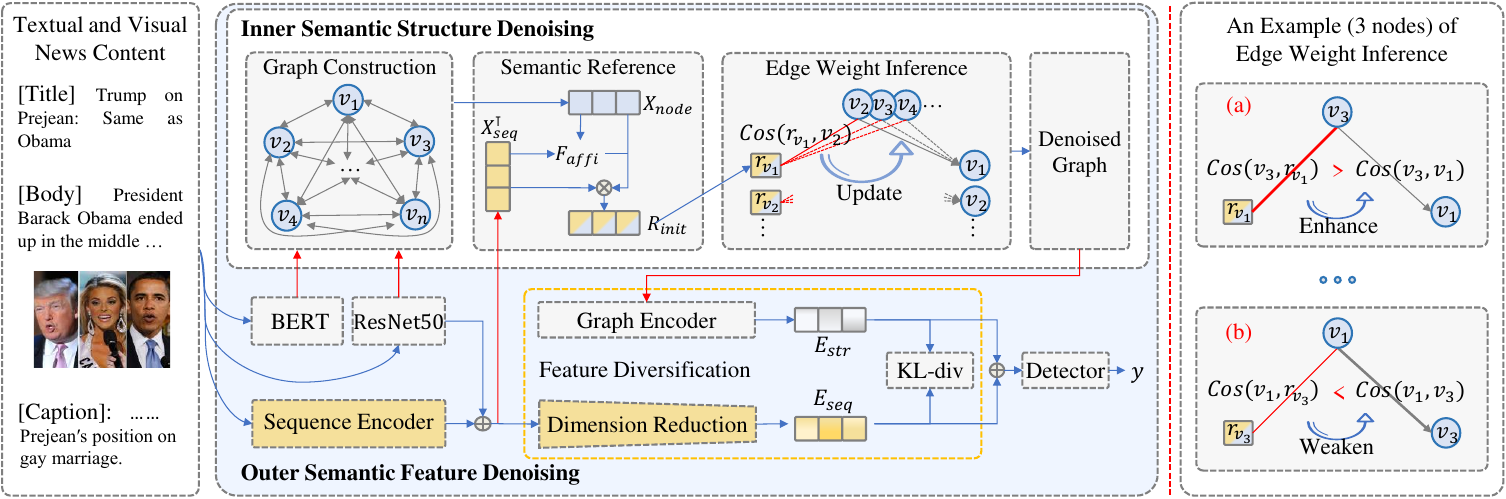}
    \caption{Overview of BREAK with Edge Weight Inference Example. The left part illustrates the overall process of BREAK, 
    where $X_{seq}$ and $E_{seq}$ denote the sequence features that used as the lower bound of semantics, $R_{init}$ indicates the reference semantics that integrate structural and sequential semantics. $X_{node}$ and $F_{affi}$ separately denote the node features and affinity matrix, and $E_{str}$ represents the structural features of the denoised graph. The right part depicts an example of edge weight inference with three nodes.
    }
    \label{fig2}
\end{figure*}

\section{Related Work}
Existing fake news detection methods focus on news content or incorporate supplementary information (e.g., comments \cite{defend}, propagation networks \cite{kdd2024propagation} and user interactions \cite{2024MMuser}). However, methods using external information often neglect in-depth news content exploration, potentially oversimplifying complex news articles into mere nodes within social networks \cite{Decor}. 
This simplification is particularly problematic during the initial stages of news dissemination, where the extra data can be both insufficient and irrelevant \cite{ALGM}\cite{2023Find}.
Meanwhile, external data like comments may contain massive noise \cite{defend}. Hence, we focus on a foundational approach to fake news detection, i.e., exploring news content.

Sequence-based methods mainly treat news text as sequence data, utilizing sequential models like recurrent neural networks (RNNs) or pre-trained language models (PLM) to capture news features. \cite{incongruent} constructs a hierarchical RNN network to capture the incongruent between the body text and news title. \cite{Metaprompt} devises a novel prompt paradigm to fully extract the semantics of news by a PLM. Moreover, \cite{IPMprogress} models multi-modal news through a progressive fusion network, which fuses cross-modal features from various levels. However, these methods fail to explicitly model the broad-range semantics hidden in complex news content \cite{22Evidence}. 

Therefore, some graph-based methods depict news content as a word-level or sentence-level graph, which explicitly represents semantics between words or sentences by an edge. \cite{22Evidence} and \cite{Disco} construct a word-level content graph based on a 2- and 3-size sliding window, respectively, while \cite{www2024msynfd} constructs a multi-hop (3- or 4-hop depending on the dataset) word-level graph to capture syntax relations. However, as outlined in Section \ref{section1}, these graphs constructed with sliding windows may neglect certain semantics. Hence, several methods construct a sentence-level graph for news content \cite{vaibhav2019} \cite{21Compare} \cite{2023TCSS}. These methods randomly initialize node features and detect fake news through the interaction patterns between sentences, which ignore the semantics of news sentences.

In the field of text classification in natural language processing, many counterparts represent text as graphs from word or sentence perspectives. However, they are also suffering from the two challenges we mentioned before. For instance, sliding window graphs \cite{NLP-sildwin} or sequence graphs (connected according to the original order of words or sentences) \cite{23EMNLP} neglect some broad-range semantics. Meanwhile, sentence-level graphs \cite{TextRank} or master-node-enhanced graphs \cite{Master} (introducing a master node to connect all others) lead to densely connected and highly-similar node features \cite{23EMNLP}.

In summary, BREAK is different from previous works in two aspects: (1) We represent news content as a more general graph structure, and any extension of existing works will confront the challenges we encountered, such as a large-sized sliding window also introducing vast noise. (2) We devise an effective network to denoise the irrelevant information and extract key sentences while only relying on news content.

\section{Approach}
BREAK aims to detect fake news solely based on news content by capturing broad-range semantics. It involves two modules under the bi-level optimization paradigm: (1) inner semantic structure denoising and (2) outer semantic feature denoising, as illustrated in Figure \ref{fig2}. Specifically, our fully connected graph naturally captures full-range semantics by modeling all potential structural information, while inevitably introducing both structural and feature noise. Therefore, the inner module utilizes the sequential semantics provided by the sequence model as a lower bound to help mitigate structural noise. Subsequently, by aligning the representations of the graph and sequence encoders in the high-dimensional space, the outer module diversifies semantic features and extracts broad-range semantics based on the denoised graph structure.

\subsection{Problem Definition}
Without loss of generality, we leverage both textual ($T$) and visual ($I$) news content for fake news detection, as they are commonly found on social media and are increasingly popular among readers \cite{IPMprogress} \cite{ALGM}. 
The textual content encompasses the news title, body text, caption, and others, providing a comprehensive representation of the news textual content. 
We formally define a piece of fake news with these elements in two forms: 
(1) The sequential semantics $E_{seq}$ obtained by a sequence encoder and image vectorization tool. 
(2) The structural semantics $E_{str}$ acquired by a graph encoder. 
Our final objective is to train a detector $f$ to classify news as real ($y\!=\!0$) or fake ($y\!=\!1$) based on $E_{seq}$ and $E_{str}$, i.e., $f(E_{seq}, E_{str}) \! \rightarrow \!y \!\in \! \{0,1\}$.

\subsection{The Overview of BREAK}
BREAK aims to achieve the following three objectives: (1) uniting structural and sequential semantics to narrow bounds and obtain a denoised graph structure; (2) diversifying semantic features to realize feature denoising while integrating broad-range semantics; and (3) enabling effective fake news detection.

The structure denoising process depends on the semantic features, and a well-learned semantic feature is determined by the denoised graph structure, indicating a close and mutual influence between them. Therefore, we introduce bi-level optimization to iteratively optimize these two denoising processes. For a piece of news, BREAK first acquires its sequential semantics from the sequence encoder and then represents the news content as a bidirectional fully connected graph to capture all potential semantics. Accordingly, with the support of sequential semantics, BREAK learns a denoised graph structure. In the outer module, a graph encoder is introduced to capture the denoised semantic interrelations and integrate them with the sequential semantics as the broad-range semantics. Ultimately, BREAK performs detection based on these broad-range semantics.

Formally, we represent BREAK as a detect function $f_{\phi, \theta}(\cdot)$, where $\phi$ indicates the parameters in the inner optimization process, and $\theta$ denotes other parameters of BREAK except $\phi$. The overall framework can be outlined as follows:
\begin{align}
    \theta^* = \hspace{0.1cm} &\underset{\theta}{argmin} \hspace{0.1cm} \mathcal{L}_{outer}(f_{\phi^{*},\theta}(\cdot), Y),\\
    \textit{s.t. } \phi^* & = \underset{\phi}{argmin}  \hspace{0.1cm} \mathcal{L}_{inner}(f_{\phi,\theta}(\cdot), Y), 
\end{align}
where $\theta^*$ and $\phi^*$ are the optimal parameters, $Y \in \{0,1\}$ is the ground-truth labels of news, and $\mathcal{L}_{inner}$ and $\mathcal{L}_{outer}$ depict the loss functions used in the inner and outer levels, respectively.

\subsection{Inner Semantic Structure Denoising}
In this process, we model news sentences and images as a bidirectional fully connected graph to cover all potential semantics. Subsequently, by introducing short-range semantics from the sequence model as the lower bound, we devise a structural and sequential semantics integration mechanism to integrate semantics from both the graph and sequence. These semantics then prompt the edge weight inference mechanism to denoise the graph structure.

\textbf{Graph Construction}. We split all the news' textual ($T$) and visual ($I$) information into sentences and images, and then vectorize them respectively by pre-trained BERT \cite{Bert} and ResNet50 \cite{Resnet50}:
\begin{equation}
    X_{T} = BERT_g(T),  \hspace{0.3cm} X_{I} = ResNet50(I),
\end{equation}
% Note that we employ the fine-tuned BERT \textbf{twice} with different purposes: $BERT_1$ is used as a sequential model to obtain the sentence sequential features ($T_{seq}$) since it preserves the original order of news sentences. 
where $BERT_g$ and $ResNet50$ are utilized as learnable vectorization tools to learn the sentence ($X_{T} \in \mathbb{R}^{M \times d}$) and image ($X_{I} \in \mathbb{R}^{P \times d}$) node features, respectively. Moreover, $M$ and $P$ separately indicate the number of sentences and images.
% Meanwhile, we utilize ResNet50 only once since images are easier to distinguish from each other in the embedding space than sentences.
Subsequently, we construct a sentence-level fully connected graph, where each sentence or image corresponds to a node in the graph. Furthermore, following the previous works \cite{2021Compare} \cite{2023TCSS}, every two nodes in the graph are connected by a bi-directed edge to reflect the forward and backward context information, as depicted in Figure \ref{fig2}. Therefore, the graph can be represented as $G = (V, E)$ with the adjacency matrix $A$, edge weight matrix $W_e$, and feature matrix $X_{node} \!=\! (X_{T},\! X_{I})$, where both $A$ and $W_e$ are all-one matrix, $X_{node} \!\in \!\mathbb{R}^{N\times d}$, $N\!=\!M\!+\!P$ is the node number of $G$, and $d$ depicts the hidden dimension of node feature.

\textbf{Structural and Sequential Semantics Integration}. We aim to alleviate the noise  in $G$, i.e., enhancing the key semantics and weakening others. Specifically, to reduce the computational complexity and avoid optimizing the graph structure and features simultaneously (since it is more difficult than optimizing either aspect alone), we only modify the graph structure by adjusting the edge weights, which can also guide the aggregation process of graph features.

A fully connected graph generates massive noise, disrupting sentence order, and node features are highly similar (as shown in Figure \ref{fig1}(d)). Due to the aforementioned issues, it is challenging to adjust edge weights within the graph aggregation mechanism without manipulating node features.
However, the sequential features preserve the original sentences' order and provide clean (well-learned) sequential semantics, which can be used as a lower-bound representation of a fully connected graph to prompt the denoise process.

In particular, we employ another pre-trained BERT to acquire sentence sequential features by $S_T \!=\! BERT_s(T) \!\in \!\mathbb{R}^{M \times d}$. 
Note that we employ \textbf{two} independent pre-trained BERTs for different purposes. $BERT_g$ is utilized as a node feature generator, where sequential semantics are disrupted by a fully connected structure. In contrast, $BERT_s$ preserves the natural order of news sentences during the optimization, which provides \textbf{cleaner} sequential semantics than a fully connected graph. At last, we concatenate $S_T$ with $X_I$ to obtain the sequential content features as $X_{seq} \!=\! S_T\! \oplus \!X_I$.

Based on $X_{seq} \in \mathbb{R}^{N \times d}$, we assign an affinity matrix $F_{affi} \in \mathbb{R}^{N \times N}$ to weigh the affinity between the node and sequential content features. Accordingly, node features $R_{init} \in \mathbb{R}^{N \times d}$ that integrate the node and sequential content features (i.e., the reference semantics) can be calculated as follows:
\begin{align}
    &F_{affi} = tanh(X_{node} W_{F} X_{seq}^{\top}), \label{eq4} \\
    R_{init} = &tanh(X_{node} W_{node} + F_{affi} X_{seq} W_{seq}), \label{eq5}
\end{align}
where $W_F, W_{node}$, and $W_{seq} \in \mathbb{R}^{d \times d}$ are the weight parameters. $R_{init}$ is effective \textit{since it considers broader-range semantics from node and sequential content simultaneously}.

\textbf{Edge Weight Inference}. As depicted in the right part of Figure \ref{fig2}, the edge weight inference process is conducted based on the  $R_{init}$. 
Taking the nodes $v_1$ and $v_3$ as an example, we first calculate the affinity between them and their corresponding sequential content features by Eq. \eqref{eq4}: $f_{v_1}, f_{v_3}\! \in \!F_{affi}$. Subsequently, their initially integrated node features $r_{v_1}, r_{v_3}\! \in \!R_{init}$ are obtained by Eq. \eqref{eq5}. 
% f_{i_1} = tanh(v_{1}^{\top} w_{2} v_{seq1})

In Figure \ref{fig2}(a), $v_1$ is the target node that we need to update all its incoming edge weights, and $v_3$ is the source node that we need to justify how much information we should aggregate from it to $v_1$. 
Specifically, we treat $r_{v_1}$ as the golden representation of $v_1$ since it comprehensively integrates broad-range semantics by Eq. \eqref{eq5}. Therefore, to guide the aggregation process in the outer optimization and make $v_1$ \textbf{closer} to $r_{v_1}$, we adjust the incoming edge weights of $v_1$ according to the cosine similarity between $r_{v_1}$ and other nodes except $v_1$ (i.e., $v_2$ and $v_3$ in this toy example). 

This stems from the finding that when $Cos(v_3, \!r_{v_1})\! > \!Cos(v_3, \!v_1)$, i.e., $v_3$ is more similar with $r_{v_1}$ than with $v_1$, which indicates $v_3$ owns more golden information in $r_{v_1}$. Thus, we leverage $Cos(v_3, r_{v_1})$ to enhance the edge weight of $e_{v_3,v_1} \in E$ and thereby \textbf{\textit{aggregating more information}} from $v_3$, and finally make $v_1$ closer to $r_{v_1}$ in this indirect way. Similarly, this deduction is applicable to the cases of $Cos(v_3, r_{v_1}) \!< \!Cos(v_3, v_1)$ and $Cos(v_3, r_{v_1}) \!= \!Cos(v_3, v_1)$.  In contrast, in Figure \ref{fig2}(b), $v_3$ is the target node, $v_1$ is the source node, and all the deductions are the same as in Figure \ref{fig2}(a).

To date, by employing the structural and sequential semantic integration and the edge weight inference for \textbf{all} nodes in the graph, the inner structure denoising module has obtained the denoised graph. Meanwhile, the corresponding adjacent matrix and node features of $\hat{G}$ are represented as $\hat{A}=\hat{W_e} \cdot A$ and $\hat{X}_{node}=X_{node}$, respectively, where $\hat{W_e} \in \mathbb{R}^{N\times N}$ indicates the learned edge weights.

% Different from the previous graph prompt learning works, 

\subsection{Outer Semantic Feature Denoising}
% After the inner optimization, the edge weights can certainly aid in distinguishable node feature learning since they guide the GNN model to aggregate neighbor node features in different degrees.
Relying on the denoised graph from the inner optimization process, we can learn distinguishable node features by integrating the sequential semantics ($X_{seq}$) and the graph representations (enriched with broad-range semantics). This integration ultimately benefits the edge weight inferring process since it helps in learning better sequential semantics and node features.

\textbf{Semantic Feature Diversification}. In detail, we first obtain the graph representation $E_{str}$ by feeding the denoised graph $\hat{G}$ into a graph encoder (which is concretized as a graph convolutional network in this work). Accordingly, a two-layer multi-layer perceptron (MLP) is utilized to align the dimension of short-range semantics $X_{seq}$ with structural features $E_{str}$.
\begin{equation}
    E_{str} = GCN(\hat{G}),\hspace{0.2cm} E_{seq} = MLP_{seq}(X_{seq}),\label{eq6}
\end{equation}
where $E_{seq}$ represents the $X_{seq}$ after dimensionality reduction. 
$E_{str}, E_{seq}\in \mathbb{R}^{N\times h}$ and $h$ indicates the dimension of hidden state. Specially, $E_{str}$ is the general representation of the initially integrated braod range semantics as it aggregates all the node features.
Consequently, the $E_{str}$ is denoised and contains richer broad-range semantics than in $E_{seq}$, while the $E_{seq}$ preserves better short-range semantics since it is strictly generated according to the sentences' natural order.
Therefore, to assist in diversifying semantic features and then capturing better broad-range semantics, we bring $E_{str}$ and $E_{seq}$ closer in the high-dimensional space to \textbf{achieve alignment} by employing the KL divergence as the loss function $\mathcal{L}_{KL}$:
\begin{equation}
    \mathcal{L}_{KL} = KL(E_{str},E_{seq}). \label{eq7}
\end{equation}
% where $\mathcal{L}_{KL}$ denotes the KL loss.
% where $\mathcal{L}_{KL}$ and $KL$ denote the KL loss and KL divergence, respectively.

\textbf{Fake News Detection}. At last, we concatenate $E_{str}$ and $E_{seq}$ as the \textbf{eventual broad-range semantics} to perform fake news detection by a two-layer MLP ($MLP_{pred}$), and we leverage the binary cross-entropy function as the classify loss function as follows:
\begin{align}
    &\hat{Y} = MLP_{pred}(E_{str} \oplus E_{seq}),\\
    \mathcal{L}_{cls} = &-\sum_{i=1}^{Z} [y_ilog(\hat{y}_i)+(1-y_i)log(1-\hat{y}_i)], \label{eq9}
\end{align}
where $\hat{Y}$ is the predicted labels of news and $\hat{y}_i \in \hat{Y}$. $\mathcal{L}_{cls}$ represents the  binary cross-entropy function, $Z$ is the number of news articles, and $y_i \in Y$ indicates the ground-truth label of the news $Z_i$.

\subsection{Model Training}
\textbf{Inner-Level Training}. The final objective of the inner level is uniting the semantics from both the graph and sequence encoders, thereby learning a denoised graph structure. We optimize it by maximizing the mutual information between the news representation $E = E_{str}\oplus E_{seq}$ and the news label $Y$ to extract label-related sentences. Specifically, this mutual information can be written as:
\begin{equation}
    I(E; Y) = \sum_{E,Y} \mathbb{P}(E, Y)log\frac{\mathbb{P}(E, Y)}{\mathbb{P}(E), \mathbb{P}(Y)}.
\end{equation}
\begin{table}[]
\centering
    \caption{The statistics of the datasets.}
    \label{table1}
\resizebox{\linewidth}{!}{\begin{tabular}{ccccc}
\hline
Dataset      & GossipCop & PolitiFact & Snopes & PolitiFact-S \\ \hline
\# fake news & 2,466     & 329        & 3,177  & 1,701        \\
\# real news & 9,270     & 331        & 1,164  & 1,867        \\
\# images    & 11,736    & 298        & 0      & 0            \\  \hline
\end{tabular}}
\end{table}
Since the $\mathbb{P}(Y|E)= \mathbb{P}(E, Y)/\mathbb{P}(E)$ is intractable, inspired by the previous work \cite{22interpretable}, we introduce a parameterized variational approximation $\mathbb{P}_{\phi}(Y|E)$ for $\mathbb{P}(Y|E)$. Accordingly, we convert the maximization problem into a minimization problem with an upper bound as follows, and the detailed derivation is in the Appendix:
\begin{align}
    &\phi^* = \underset{\phi}{argmin}\hspace{0.1cm} -I(E; Y),\\
    -I(E; &Y) \leq -\mathbb{E}[\text{log}\mathbb{P}_{\phi}(Y|E)]-H(Y), \label{eq12}
\end{align}
where the $\phi^*$ is the optimal parameters of the denoise process. 
% The detailed derivation of Eq. \eqref{eq12} is listed in Appendix, Section A.1. 
% \triangleq -\Sigma_{Y}\mathbb{P}(Y)log\mathbb{P}(Y)
Meanwhile, since $H(Y)$ is the entropy of news labels $Y$ (a constant), we only optimize $-\mathbb{E}[\text{log}\mathbb{P}_{\phi}(Y|E)]$ in the inner level training process.
Furthermore, $\mathbb{P}_{\phi}(Y|E)$ is essentially work as a predictor with the parameter $\phi$. Specifically, for a piece of news $Z_i$, BREAK obtains its representation $E_i \in E$ by Eq. \eqref{eq6} and performs a prediction: $f_{\phi, \theta}(Z_i) \rightarrow Y_i$. Therefore, $\mathbb{P}_{\phi}(Y_i|E_i)$ is equivalent to the possibility of news $Z_i$ be predicted as $Y_i$ by $f_{\phi, \theta}(Z_i)$. Note that $\theta$ is frozen during the inner-level training process, and only the change on $\phi$ leads to a change on $\mathbb{P}_{\phi}(Y_i|E_i)$.
% Note that $\theta$ are frozen during the inner-level training process, $\mathbb{P}_{\phi}(Y_i|E_i)$ is only influenced by the parameters $\phi$. 
Ultimately, by expanding $-\mathbb{E}[\text{log}\mathbb{P}_{\phi}(Y|E)]$, the loss function of the inner level can be written as a standard cross entropy loss:
\begin{equation}
    \mathcal{L}_{inner} = -\sum_{i=1}^{Z} [y_ilog(\hat{y}_i)+(1-y_i)log(1-\hat{y}_i)].
\end{equation}
% Detailed proof can be found in the supplementary material.

\textbf{Outer-Level Training}. In the outer-level training step, our final goal is to diversify semantic features and learn a comprehensive news representation, which captures the broad-range semantics hidden in the news content. Therefore, we combine Eq. \eqref{eq7} and Eq. \eqref{eq9} as the final loss function of the outer level.
\begin{equation}
    \mathcal{L}_{outer} =  \mathcal{L}_{cls}+ \beta \mathcal{L}_{KL},
\end{equation}
where $\beta$ is a hyperparameter used to decide how close the $E_{str}$ and $E_{seq}$ should be.

% \subsection{Detailed Derivation of the Inner-level Training}
% \subsection*{A.1. Detailed Derivation of Eq. (12)}
\section{Experiments}
This section evaluates the effectiveness of BREAK on four datasets and aims to answer the following research questions.

\textbf{RQ1}: How does BREAK compare to baselines in fake news detection?
\textbf{RQ2}: How well does BREAK generalize in scenarios when clear evidence is available?
\textbf{RQ3}: Is every component of BREAK essential for fake news detection?
\textbf{RQ4}: How does hyperparameter $\beta$ influence BREAK's performance?
\textbf{RQ5}: Can BREAK effectively denoise the fully connected graph and capture crucial semantics?

\subsection{Datasets}
We perform fake news detection on four real-world datasets as follows to assess the detection performance of BREAK.

\textbf{Content-Only Datasets}. FakeNewsNet \cite{shu2020fakenewsnet} comprises two datasets: GossipCop and PolitiFact. Both datasets are sourced from fact-checking websites and labeled as either fake or real. Each news article includes titles and body text, and some include images.

\textbf{Clear Evidence Available Datasets}. To assess the generalization ability of BREAK in the scenario where clear evidence (one type of external data) is available, we additionally employ two datasets containing evidence: Snopes \cite{snopes} and PolitiFact \cite{politifact-s} (renamed as PolitiFact-S for differentiation). Each entry in these datasets consists of a brief news and its corresponding evidence, and the preprocessed version from \cite{22Evidence} is utilized for a fair comparison.
The detailed statistics of these datasets are in Table \ref{table1}.

\begin{table}[]
\centering
    \caption{Performance comparison of different methods on GossipCop and PolitiFact datasets, with the best performances in bold and the runners-up underlined. GET is not involved because it cannot work without evidence.}
\label{table2}
\resizebox{\linewidth}{!}{
\begin{tabular}{cccccc}
\hline
Dataset                    & Methods     & Acc. & Prec. & Rec. & F1 \\ \hline
\multirow{13}{*}{GossipCop} & BERT & 0.831 & 0.821 & 0.831 & 0.792 \\
            & GAT+2 Attn Heads & 0.846 & 0.835 & 0.846 & 0.840 \\
                           & SAFE        & 0.819 & 0.817 & 0.819 & 0.818   \\
                           & HMCAN       & 0.836 & 0.825 & 0.836 & 0.825 \\
                           & CAFE        & 0.814 & 0.824 & 0.814 & 0.819  \\
                           & MRML        & 0.817 & 0.838 & 0.817 & 0.821 \\
                           & ALGM        & 0.829 & 0.812 & 0.829 & 0.811 \\ 
                           & CSFND        & 0.835 & 0.849 & 0.835 & 0.847 \\  
                           & L-Defense   & 0.841 & 0.826 & 0.841 & 0.827 \\ 
                           \cline{2-6} 
                           & ChatGLM2-6B & 0.856 & 0.847 & 0.856 & 0.850 \\
                           & LLaMA2-7B & \underline{0.866} & \underline{0.858} & \underline{0.860} & \underline{0.860} \\ \cline{2-6} 
                           & BREAK &\textbf{0.882} &\textbf{0.876} & \textbf{0.882} & \textbf{0.871} \\ \cline{2-6} 
                           & Improve(\%) & 1.848 & 2.098 & 1.848 & 1.279 \\ \hline
                           
\multirow{13}{*}{PolitiFact} & BERT & 0.791 & 0.794 & 0.791 & 0.790 \\
& GAT+2 Attn Heads & 0.892 & 0.893 & 0.893 &  0.892\\
                           & SAFE        & 0.853 & 0.814 & 0.875 & 0.843 \\
                           & HMCAN       & \underline{0.924} & \underline{0.927} & \underline{0.924} & 0.923 \\
                           & CAFE        & 0.791 & 0.800 & 0.791 & 0.793 \\
                           & MRML        & 0.817 & 0.838 & 0.817 & 0.821 \\
                           & ALGM        & 0.887 & 0.900 & 0.887 & 0.888 \\
                           & CSFND        & 0.917 & 0.917 & 0.917 & \underline{0.929} \\ 
                           & L-Defense   & 0.879 & 0.900 & 0.879 & 0.878 \\
                           \cline{2-6} 
                           & ChatGLM2-6B & 0.892 & 0.902 & 0.892 & 0.892 \\ 
                           & LLaMA2-7B   & 0.908 & 0.914 & 0.908 & 0.908 \\ \cline{2-6} 
                           & BREAK   & \textbf{0.955} & \textbf{0.956} & \textbf{0.955} & \textbf{0.955} \\ \cline{2-6} 
                           & Improve(\%) & 3.355 & 3.128 & 3.355 & 2.799 \\ \hline
\end{tabular}}
\end{table}

\begin{table}[]
\centering
    \caption{The generalization performance of BREAK on datasets with evidence: Snopes and PolitiFact-S. The best performances are in bold, and the runners-up are underlined.}
\label{table3_}
\resizebox{\linewidth}{!}{
\begin{tabular}{cccccc}
\hline
Dataset                    & Methods     & Acc. & Prec. & Rec. & F1 \\ \hline                           
\multirow{6}{*}{Snopes}       & BERT & 0.766 & 0.753 & 0.766 & 0.741 \\
& GET         & 0.835 & 0.838 & 0.835 & \underline{0.836} \\
                              & L-Defense  & 0.766 & 0.806 & 0.774 & 0.758 \\
                              & ChatGLM2-6B & \underline{0.843} & \underline{0.851} & \underline{0.843} & 0.829 \\
                              & LLaMA2-7B   & 0.836 & 0.834 & 0.836 & 0.827 \\ \cline{2-6} 
                              & BREAK      & \textbf{0.860} & \textbf{0.858} & \textbf{0.860} & \textbf{0.859} \\ \hline
                              & Improve(\%) & 2.017 & 0.823 & 2.017 & 2.751 \\ \hline
                              
\multirow{6}{*}{PolitiFact-S} & BERT & 0.565 & 0.563 & 0.563 & 0.543 \\
                & GET   & 0.682 & 0.684 & 0.682 & \underline{0.679} \\
                              & L-Defense  & 0.632 & 0.650 & 0.625 & 0.611 \\
                              & ChatGLM2-6B & 0.676 & 0.675 & 0.676 & 0.675 \\
                              & LLaMA2-7B   & \textbf{0.710} & \textbf{0.711} & \textbf{0.710} & \textbf{0.708} \\ \cline{2-6} 
                              & BREAK      & \underline{0.709} & \underline{0.708} & \underline{0.709} & \textbf{0.708} \\ \hline
                              & Improve(\%) & -0.141 & -0.422 & -0.141 & 0.000 \\ \hline
\end{tabular}}
\end{table}

\subsection{Baselines}
We compare BREAK with twelve representative baselines:
\begin{itemize}
    \item \textbf{BERT} \cite{Bert} is a pre-trained language model, and we fine-tune its last two layers for fake news detection.
    % In this paper, BERT is utilized to simulate the performance of fake news detection when only sequential features are available.
    \item \textbf{GAT+2 Attn Heads} \cite{vaibhav2019} represents news sentences as a fully connected graph with randomly initialized node features to detect fake news.
    \item \textbf{SAFE} \cite{SAFE} detects fake news by exploring the similarity between textual and visual information.
    \item \textbf{HMCAN} \cite{HMCAN} devises a hierarchical multi-modal attention network to learn a multi-modal news representation.
    \item \textbf{CAFE} \cite{CAFE} aligns cross-modal features and detects fake news by cross-modal ambiguity.
    \item \textbf{GET} \cite{22Evidence} treats news and evidence as word-level graphs, respectively. Capturing the long-range semantics to improve fake news detection.
    \item \textbf{MRML} \cite{mrml} extracts multi-modal relationships and detects multi-modal rumors based on deep metric learning.
    \item \textbf{ALGM} \cite{ALGM} proposes a framework based on the Markov random field and fuses cross-modal features by ambiguity.
    \item \textbf{CSFND} \cite{CSFND} devises an unsupervised fake news detection framework to capture the relationships between news semantic feature space and fake news decision space.
    \item \textbf{L-Defense} \cite{L-Defense} leverages the wisdom of crowds to extract evidence and generate justifications via prompting LLM.
    \item \textbf{ChatGLM2-6B} \cite{ChatGLM2} and \textbf{LLaMA2-7B} \cite{llama2} are two large language models (LLMs) with 6B and 7B parameters, respectively. We utilize the same training datasets to \textbf{fine-tune} them (using Lora \cite{lora}) as the baselines for comparison.
\end{itemize}
Where GAT+2 Attn Heads, GET, and ALGM leverage the GNN model to improve fake news detection. Meanwhile, LLMs are theoretically suitable for long text modeling since they are trained to handle long sequences of tokens with billions of parameters. In other words, LLMs are strong and competitive baselines, but they come with higher fine-tuning costs and require more advanced fine-tuning skills compared to traditional methods.

\begin{figure}[htbp]
    \centering
    \includegraphics[width=\linewidth,height=0.33\textwidth]{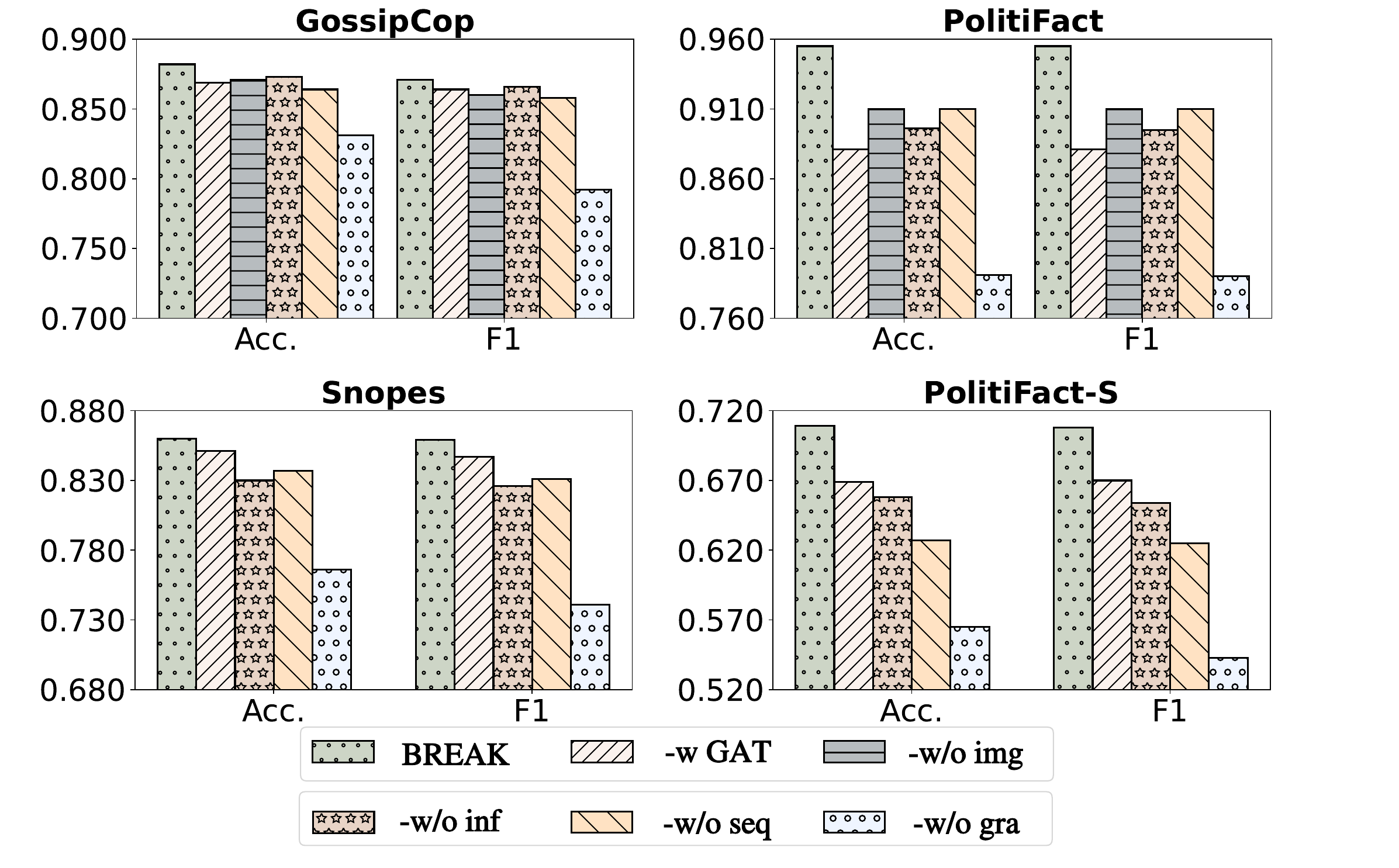}
    \caption{Ablation study on four datasets.}
    \label{fig3}
\end{figure}

% \subsection{Implementation Details}
% We partition each dataset into training, validation, and testing sets in an 8:1:1 ratio. We fine-tune the last two layers of $BERT_s$ and $BERT_g$ and the last fully connected layer of $ResNet50$. For the GossipCop, PolitiFact, Snopes, and PolitiFact-S datasets, BREAK is trained in the batch size of 8, 8, 64, and 64 with a hyperparameter $\beta$ of 0.1, respectively. The learning rates of inner and outer levels are set as 0.1 and 0.00001 separately. Moreover, the dimensions of $d$ and $h$ are set as 768 and 128, respectively. The early-stopping patience is determined to be 8, and Adam \cite{Adam} is employed as the optimizer. For the metrics, we utilize the weighted accuracy (Acc.), precision (Prec.), recall (Rec.), and F1 score to alleviate the influence of unbalanced datasets.
% \usepackage{multirow}

\subsection{Performance Comparison (RQ1)}
To evaluate the performance of BREAK on content-only (early) fake news detection, we compare it with eleven advanced baselines on the GossipCop and PolitiFact datasets, as shown in Table \ref{table2}.

BREAK achieves the highest performance across all metrics, leading to a notable improvement of 3.69\% and 3.47\% in F1 scores compared to sub-optimal results (excluding LLMs) on the GossipCop and PolitiFact datasets, respectively. This improvement underscores the effectiveness of BREAK in fully exploring news content.

Moreover, among all \textit{traditional} baselines, the ``GAT+2 Attn Heads'' method exhibits competitive results on the GossipCop dataset and the PolitiFact dataset. This finding validates the effectiveness of sentence-level graph construction. However, ``GAT+2 Attn Heads'' falls short when competing with BREAK and novel content-based methods like CSFND, as it initializes node features randomly, overlooking sentence semantics and potential graph noise during the optimization process. 
Meanwhile, transformer-based methods like HMCAN, MRML, and CSFND focus on short-range semantics while neglecting some long-range semantics of text. CAFE performs significantly worse than other baselines on PolitiFact due to the insufficient images in PolitiFact. In contrast, ALGM further incorporates Markov random fields and semi-supervised settings to improve detection performance. L-Defense shows poorer performance compared with LLMs as it doesn't fine-tune the involved LLM.
For LLM baselines, LLaMA2-7B exhibits superior detection results compared to ChatGLM2-6B. We attribute this improvement to the fact that LLaMA2-7B is trained on a larger English corpus than ChatGLM2-6B. Even though LLMs show promising detection performance, they require more fine-tuning costs and are unstable across various datasets. However, our BREAK outperforms them to varying degrees, relying only on a much smaller pre-trained language model, BERT, and a stable hyperparameter across various datasets (details about the hyperparameter can be found in Section \ref{4.7}).

\begin{figure}[htbp]
    \centering
    \includegraphics[width=\linewidth,height=0.23\textwidth]{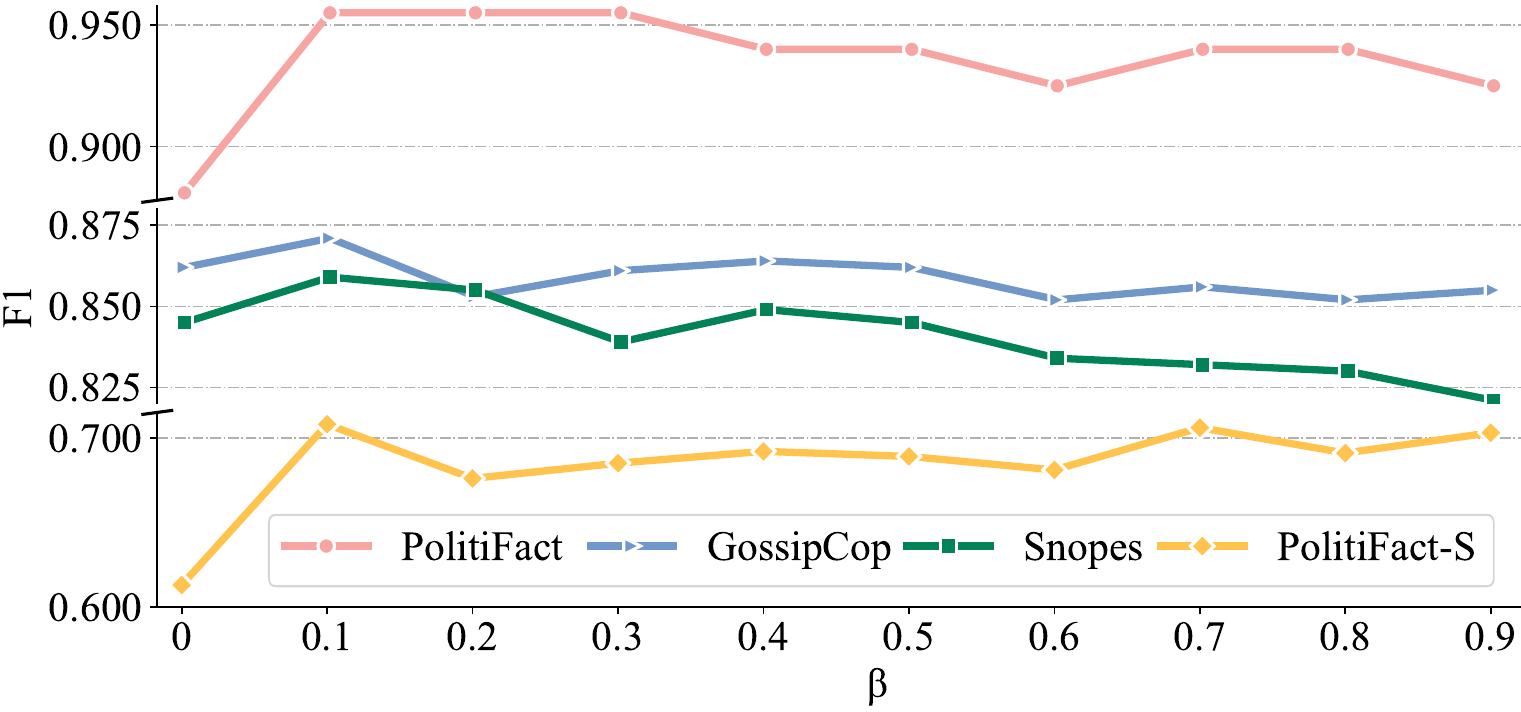}
    \caption{Hyperparameters sensitivity with regard to $\beta$.}
    \label{fig4}
\end{figure}
\subsection{Generalization Exploration (RQ2)}
In a real application, \textbf{clear} evidence sometimes is available, e.g., the official announcement about the news event. Therefore, we investigate the detection performance of BREAK when such evidence is accessible. Specifically, we conduct experiments on Snopes and PolitiFact-S datasets, in which the evidence directly pertains to the news under consideration. The results are outlined in Table \ref{table3_}.

We compare BREAK with BERT and four strong baselines: the state-of-the-art method (GET) and three methods with LLMs. The results in Table \ref{table3_} demonstrate that our BREAK exhibits prominent performance, boosting the F1 score by 2.75\% compared to the runners-up results on Snopes, and drawing near the detection results produced by LLaMA2-7B on PolitiFact-S. Particularly, GET outperforms some baselines to various degrees in F1 score, benefiting from the long-range semantics captured by its word-level graph. However, the word-level graph used in GET overlooks some semantics we discussed in Section \ref{section1}, while BREAK models and denoises all semantics appropriately. 
Therefore, the results on four datasets demonstrate BREAK's generality and stability. Moreover, these results proved that BREAK is effective for both long (GossipCop and PolitiFact) and short (Snopes and PolitiFact-S) content.

\subsection{Ablation Study (RQ3)}
To assess the necessity of each part in BREAK, we compare BREAK with its five variants: ``-w GAT,'' ``-w/o img,'' ``-w/o inf,'' ``-w/o seq,'' and ``-w/o gra.'' Specifically, ``-w GAT'' is the variant that uses graph attention network (GAT) as the denoise module and graph encoder, ``-w/o img'' excludes visual content in the GossipCop and PolitiFact datasets, ``-w/o inf'' omits the sequential semantic integration mechanism and edge weight inference mechanism, ``-w/o seq'' omits the sequence encoder and only employing GCN for fake news detection (i.e., sequential semantics), and ``-w/o gra'' means without the graph structure and only use BERT as the detection model. 

The comparison results in Figure \ref{fig3} show that the absence of any part of BREAK leads to sub-optimal performance. In detail, ``-w GAT'' shows that the attention mechanism cannot handle such complex semantic interrelations well. Moreover, the performances of "-w/o inf" on all datasets are very close those of "-w/o seq," as "-w/o inf" considers noise semantics during semantic integration. These experimental phenomena demonstrate the effectiveness of the inner denoising process. Meanwhile, both ``-w/o seq'' and ``-w/o gra'' show sub-optimal results, with ``-w/o gra'' performing the worst, indicating that integrating broad-range semantics from the graph and sequence is necessary, and that structural semantics are more crucial for comprehensive news representation modeling.

\begin{figure}[!t]

\centering
\subfloat[The visualization of sentences' weights.]{\includegraphics[width=8cm]{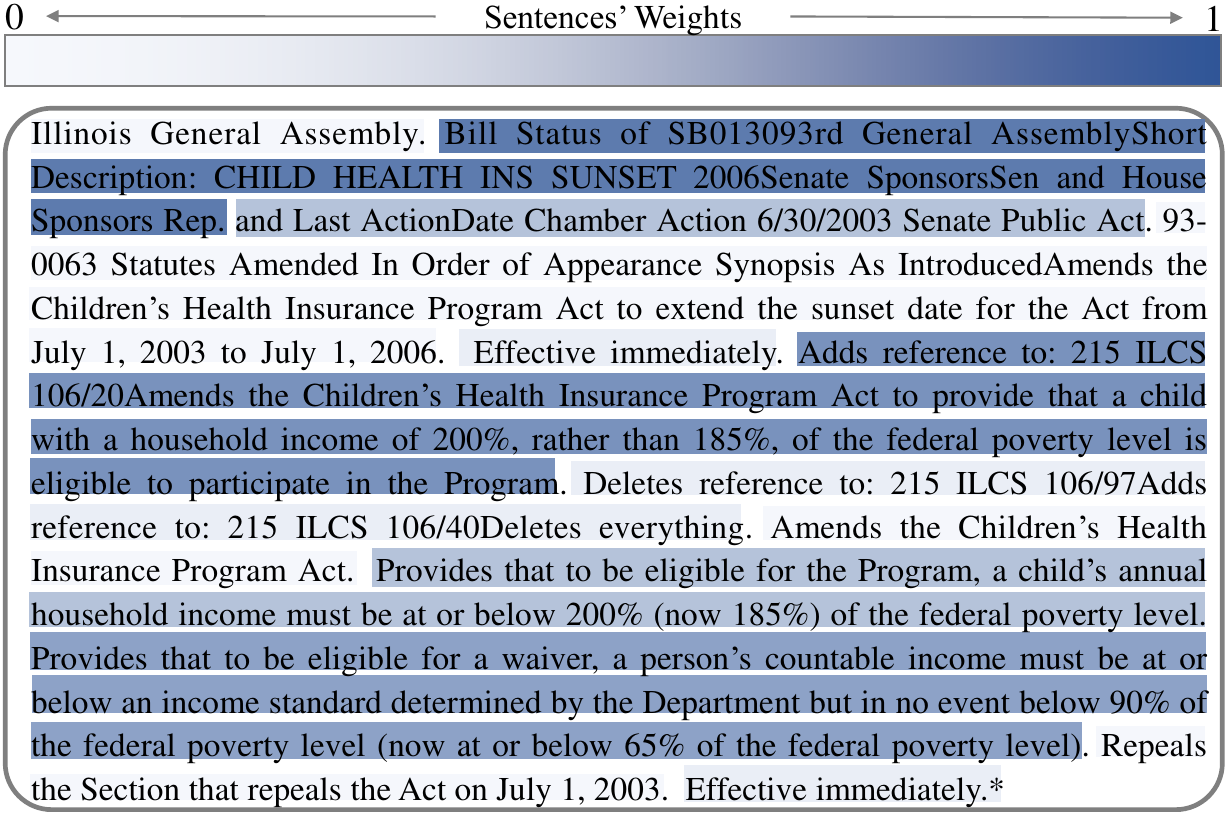}%
\label{fig5a}}
\hfil
\subfloat[Denoised edge weights.]{\includegraphics[width=4cm]{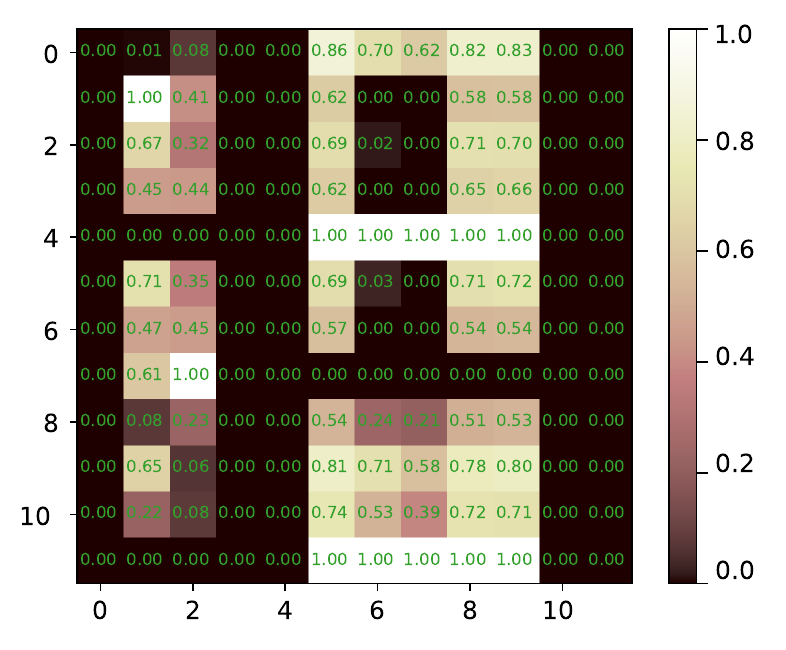}%
\label{fig5b}}
\hfil
\subfloat[Degrees of nodes.]{\includegraphics[width=4cm]{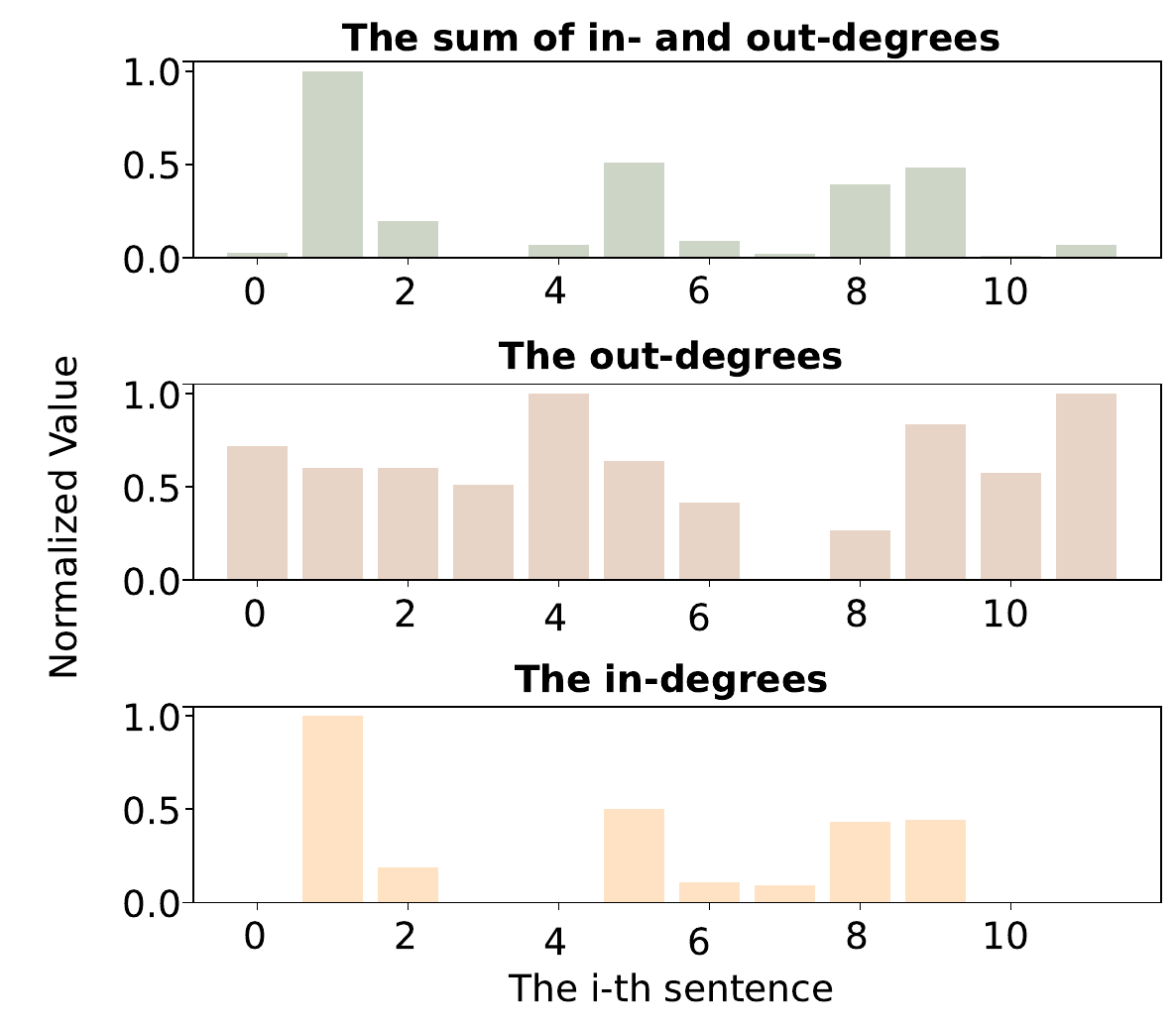}%
\label{fig5c}}
\caption{A case study of structure denoising. (a) represents the visualization of sentences' weights. (b) represents the weight of each edge between sentences. (c) depicts the normalized in- and out-degrees of each node.}
\label{fig5}
% \vspace{-5pt}
\end{figure}

\subsection{Sensitivity of Hyperparameter $\beta$ (RQ4)}
\label{4.7}
In the BREAK model, the hyperparameter $\beta$ balances the combination of structural and sequential semantics. We analyze its impact on fake news detection performance, with results shown in Figure \ref{fig4}. BREAK \textit{\textbf{consistently}} achieves optimal results across four datasets with $\beta=0.1$, demonstrating its necessity and generalizability. Meanwhile, significant improvements are observed in both PolitiFact and PolitiFact-S, consistently surpassing results with $\beta$ values from 0.1 to 0.9 compared to $\beta=0$. However, larger $\beta$ values lead to various degrees of performance degradations on all datasets, which incorporate \textit{more noise} from the fully connected graph to short-range semantics and thereby hinder the denoise process.

\subsection{Case Study on Structure Denoising and Feature Diversification (RQ5)}
\begin{figure}[!t]
\centering
\subfloat[Original similarity]{\includegraphics[width=4cm]{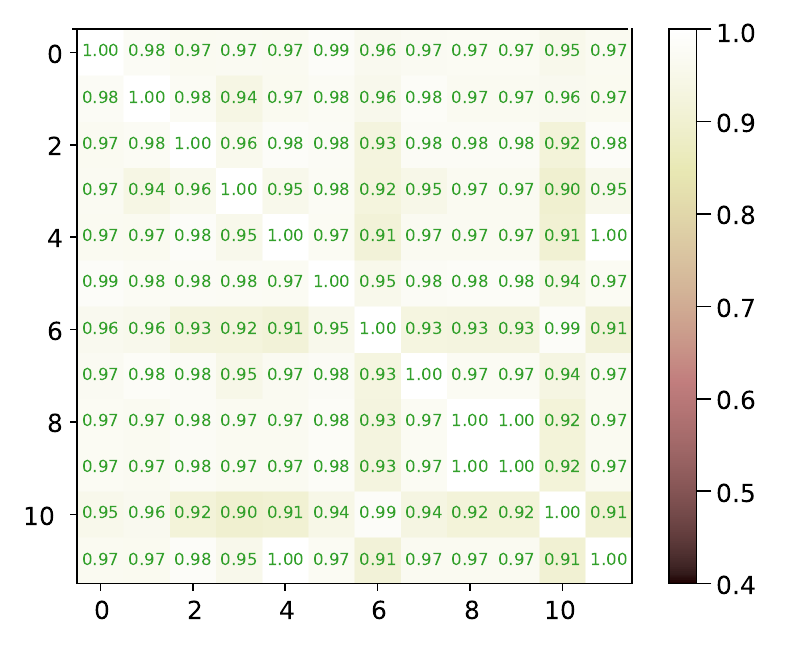}%
\label{fig6a}}
\hfil
\subfloat[Diversified similarity]{\includegraphics[width=4cm]{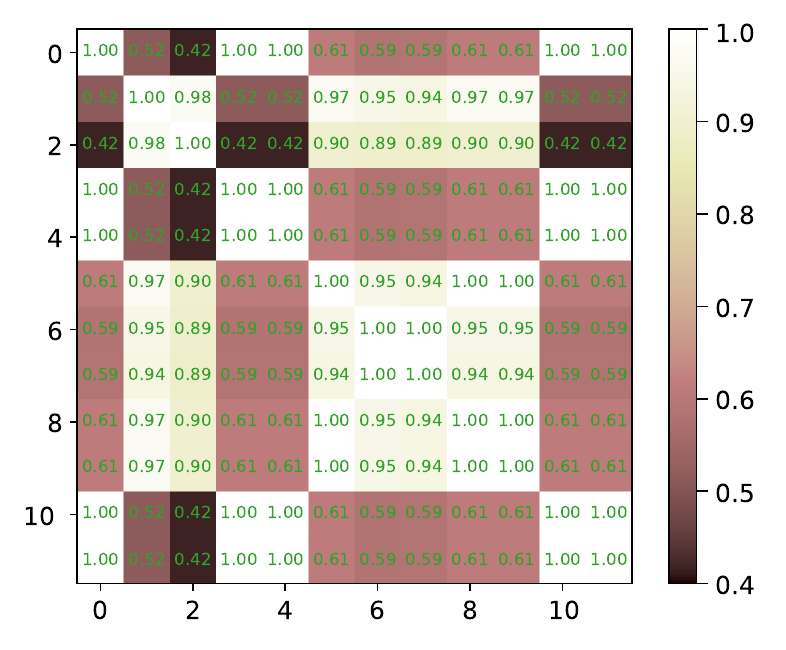}%
\label{fig6b}}
\caption{A case study of node features diversification. (a) indicates the original similarity of news sentences. (b) denotes the similarity of news sentences learned by BREAK.}
\label{fig6}
\vspace{-5pt}
\end{figure}

We evaluate the effectiveness of BREAK in structure denoising and feature diversification through a case study. Specifically, we visualize the edge weights of graph and the similarity between sentences (node features), as depicted in Figure \ref{fig5} and Figure \ref{fig6}. 
% The news text of the case study is provided in Section A.2 of the Appendix.

\textbf{Structure Denoising}. The edge weights learned by BREAK in two directions are illustrated in Figure 5\subref{fig5b}, where rows (columns) indicate the forward (backward) direction. In BREAK, the noise semantics are weakened to close to 0, and the key semantics are enhanced to around 1. Moreover, these edge weights also exhibit certain interpretability for the detection results. Specifically, we sum all the out-degrees and in-degrees of a node (sentence) as its weight, and the normalized results are shown in Figure 5\subref{fig5c}. Surprisingly, we observe that node weights are primarily determined by in-degrees (backward connections). 
% Additionally, a majority of nodes disseminate their information to other nodes, exhibiting high out-degrees. From this observation, we can conclude that sentences with high in-degrees are likely to be topic sentences, as other sentences revolve around these focal points.
Meanwhile, as depicted in Figure 5\subref{fig5a}, BREAK assigns distinct weights to sentences, revealing key facts crucial for fake news detection. In detail, sentences with higher weights convey the facts that the proposal of a children's health bill and the resulting modifications in eligibility criteria. 
% Moreover, we observe that sentences with higher weights predominantly function as factual statements, resulting in higher in-degrees. Conversely, sentences with lower weights play a supporting role, leading to higher out-degrees. It's noteworthy that news articles tend to reiterate facts, such as ``effective immediately'' and ``change from 185\% to 200\%'', contributing to the high similarity among sentences.

\textbf{Feature Diversification}. The cosine similarity between news sentences pre- and post-training is presented in Figure 6\subref{fig6a} and Figure 6\subref{fig6b}. The initial similarity obtained by the BERT model is notably high, with values consistently exceeding 0.9. In contrast, the sentence features learned by our BREAK are more distinct. This can effectively mitigate the over-smoothing problem during graph aggregation and aid in identifying key news sentences.

\section{Conclusion}
In this paper, we propose BREAK, a network devised to extract and integrate the broad-range semantics while avoiding noise incorporation. BREAK models broad-range semantics as a fully connected graph and implements dual denoising modules under a bi-level optimization paradigm to mitigate both structural and feature noise. These two modules effectively eliminate irrelevant semantic interrelations and diversify semantic features, respectively. Specifically, at the inner level, we introduce a sequence-based structure to obtain the sequential semantics and the lower bound of structure. Moreover, a structural and sequential semantic integration mechanism and an edge weight inference mechanism are devised to achieve structural denoising by narrowing the structure bounds. At the outer level, we employ KL-divergence to align the graph and sequence encoders, thereby diversifying semantic features and integrating broad-range semantics for detection. Extensive experiments on both content-only and clear-evidence-accessible scenarios demonstrate the superiority of BREAK in fake news detection.

\begin{acks}
This work is supported by the National Natural Science Foundation of China (No. 62176028 and 62302417) and the Science and Technology Research Program of Chongqing Municipal Education Commission (No. KJZD-K202204402 and KJZD-K202304401). Coauthor Kai Shu consulted on this paper on unpaid weekends for personal interest.
\end{acks}
% \end{verbatim}

% \section{Appendices}

% If your work needs an appendix, add it before the
% ``\verb|\end{document}|'' command at the conclusion of your source
% document.

% Start the appendix with the ``\verb|appendix|'' command:
% \begin{verbatim}
%   \appendix
% \end{verbatim}
% and note that in the appendix, sections are lettered, not
% numbered. This document has two appendices, demonstrating the section
% and subsection identification method.


\bibliographystyle{ACM-Reference-Format}
\balance
\bibliography{sample-base}

\appendix

\section{Detailed Derivation of Inner-Level Training}

\subsection{Prefix Knowledge}
Some prefix knowledge involved in the derivation is as follows:

\textbf{Self Information}. Given a random variable $a \in \mathcal{A}$. Its self information can be written as $S(a) = -log \hspace{0.1cm} \mathbb{P}(a)$, where $\mathbb{P}$ indicates the distribution of $a$.
    
\textbf{Entropy}. The entropy of $a$ is defined as the expectation of $S(a)$: $H(a) = \mathbb{E}_a[S(a)] = -\sum_{a\in \mathcal{A}} \mathbb{P}(a)log\hspace{0.1cm} \mathbb{P}(a)$.
    
\textbf{KL Divergence}. KL divergence measures the discrepancies between two distributions. Specifically, given random variable $a$ and its two distributions (true and predicted) $\mathbb{P}(a)$ and $\mathbb{Q}(a)$, the KL divergence can be written as $KL(\mathbb{P}(a),\mathbb{Q}(a))=\sum_{a\in \mathcal{A}}\mathbb{P}(a)\frac{\mathbb{P}(a)}{\mathbb{Q}(a)}$, which can be utilized to measure how the predicted distribution $\mathbb{Q}(a)$ close to the true distribution $\mathbb{P}(a)$.
    
\textbf{Mutual Information}. The mutual information $I(a;b)$ quantifies the reduction in uncertainty about variable $a$ when the value of another variable $b$ is known, i.e., assesses the degree of dependence or correlation between $a$ and $b$.
    Formally, the mutual information can be described as $I(a;b)=\sum_{a\in \mathcal{A}}\sum_{b \in \mathcal{B}}\mathbb{P}(a,b)\log \frac{\mathbb{P}(a,b)}{\mathbb{P}(a) \mathbb{P}(b)}$. 
% \end{itemize}

\subsection{Detailed Derivation}
The inner structure denoising process aims to extract key sentences or images for effective fake news detection. Utilizing known news labels in the training set, we seek to maximize the mutual information between news representations and labels. This maximization assists the denoising process in excavating the most label-related information and eliminating non-relevant noise.

By the definition of mutual information, we can formally write this optimization objective as Eq.(10).
Following the previous work \cite{22interpretable}, we give the detailed derivation of transforming Eq. (10) into Eq. (12).
BREAK detects fake news by the representation of news $E$, i.e., $\mathbb{P}(Y|E)=\frac{\mathbb{P}(E, Y)}{\mathbb{P}(E)}$. However, $\mathbb{P}(Y|E)$ is intractable since no ground-truth label to model the relationship between $E$ and $Y$. Therefore, we introduce a variational approximation $\mathbb{P}_{\phi}(Y|E)$ for $\mathbb{P}(Y|E)$. Accordingly, we acquire the lower bound of Eq. (10) as follows:
\begin{align} 
    &I(E; Y) = \sum_{E, Y} \mathbb{P}(E, Y)log \hspace{0.1cm}\frac{\mathbb{P}(Y|E)}{\mathbb{P}(Y)}\\
    &=\mathbb{E}_{E,Y} \left [ log \hspace{0.1cm} \frac{\mathbb{P}(Y|E)}{\mathbb{P}(Y)} \right ] \\
    &=\mathbb{E}_{E,Y} \left [ log \hspace{0.1cm} \frac{\mathbb{P}_{\phi}(Y|E)}{\mathbb{P}(Y)} \right ] +\mathbb{E}_{E,Y} \left [ KL(\mathbb{P}(Y|E),\mathbb{P}_{\phi}(Y|E)) \right ] \label{derivation} \\
    &\geq \mathbb{E}_{E,Y} \left [ log \hspace{0.1cm} \frac{\mathbb{P}_{\phi}(Y|E)}{\mathbb{P}(Y)} \right ] \\
    &=\mathbb{E}_{E,Y} \left [ log \hspace{0.1cm} \mathbb{P}_{\phi}(Y|E) \right ] - \mathbb{E}_{E,Y}\left [ log \hspace{0.1cm} \mathbb{P}(Y) \right ] \\
    &=\mathbb{E}_{E,Y} \left [ log \hspace{0.1cm} \mathbb{P}_{\phi}(Y|E) \right ] + H(Y),
\end{align}
where the KL divergence in step \eqref{derivation} is utilized to measure the difference between the true distribution $\mathbb{P}(Y|E)$ and the variational approximation $\mathbb{P}_{\phi}(Y|E)$. Ultimately, we acquire Eq. (12) by inverse the lower bound as $-I(E; Y) \leq -\mathbb{E}_{E, Y} \left [ log \hspace{0.1cm} \mathbb{P}_{\phi}(Y|E) \right ] - H(Y)$.

\section{Implementation Details}
We partition each dataset into training, validation, and testing sets in an 8:1:1 ratio. We fine-tune the last two layers of $BERT_s$ and $BERT_g$ and the last fully connected layer of $ResNet50$. For the GossipCop, PolitiFact, Snopes, and PolitiFact-S datasets, BREAK is trained in the batch size of 8, 8, 64, and 64 with a hyperparameter $\beta$ of 0.1, respectively. The learning rates of inner and outer levels are set as 0.1 and 0.00001 separately. Moreover, the dimensions of $d$ and $h$ are set as 768 and 128, respectively. The early-stopping patience is determined to be 8, and Adam \cite{Adam} is employed as the optimizer. For the metrics, we utilize the weighted accuracy (Acc.), precision (Prec.), recall (Rec.), and F1 score to alleviate the influence of unbalanced datasets.

\section{More Analyses of the Experiments}
\subsection{Generalization Exploration (RQ2)}
In this subsection, we further analyze the experiments and the advantages of BREAK to prove the generalization of BREAK on news length. 

\textbf{Analysis}.We conducted experiments on both long and short news datasets, the results of which demonstrate the generalization of BREAK on news length. Specifically, the news in the GossipCop and PolitiFact datasets often consists of dozens of sentences, while the news in Snopes and PolitiFact-S is typically a single sentence, with the remainder consisting of related evidence. Based on the experimental results, BREAK performs well on all four datasets, suggesting that news length will not influence the superiority of BREAK.

\textbf{Advantage}. The reason why BREAK can adapt to different news lengths is that we integrate BERT and GNN models. With this architecture, longer news can be accommodated by increasing BERT's input length, with the text truncated or zero-padded as needed. The zero-padded sections are excluded from updates. For graph construction, only non-zero-padded sentences are used as nodes, and since the GNN model can handle graphs with varying node counts, the model's inference is unaffected by differences in the number of sentences in the news.

\subsection{Ablation Study (RQ3)}
In this subsection, we list more experiments about the ablation of KL-divergence. Based on analysis and experimental results, we employ KL-divergence to achieve feature alignment. In detail, we compare KL-divergence with cosine similarity (Cos) and Euclidean distance (Euc), and the analysis and experimental results are as follows:

\textbf{Analysis}. KL-divergence measures the distance between probability distributions, which reflects sentence importance. Meanwhile, using sequence features as target distributions provides a clear reference for learning structure features, unlike cosine similarity.

\textbf{Results}. To more intuitively demonstrate the differences of the above three measurement methods, we substitute Cos (-w Cos) and Euc (-w Euc) for KL-divergence respectively. As a result, the experimental results of the F1 score from the four data sets are shown in Table \ref{table4}. We can observe that BREAK achieves the best performance, which is consistent with the analysis and demonstrates the advantage of using KL-divergence.

\textbf{Sources of Performance Gains}. The performance improvements in BREAK are primarily attributed to its dual denoising mechanism, which balances noise removal and the preservation of broad-range label-related information. This iterative mechanism refines both structural and sequential semantics, ensuring robust semantic integration. This is evidenced by the empirical validation (i.e., ablation study). In detail, ablation experiments (e.g., "-w GAT" and "-w/o inf") demonstrate the necessity of dual-level optimization. Gradual refinements to structure and feature representations enable the model to extract meaningful interrelations while mitigating the impact of noise, underscoring the importance of this mechanism.

\begin{table}[]
\centering
    \caption{The ablation study of KL-divergence on four datasets.}
\label{table4}
\resizebox{\linewidth}{!}{
\begin{tabular}{ccccc}
\hline
Methods       & GossipCop & PolitiFact & Snopes & PolitiFact-S\\ \hline                           
BREAK  & \textbf{0.871} & \textbf{0.955} & \textbf{0.859} & \textbf{0.708} \\                       
-w Cos & 0.842 & 0.926 & 0.850 & 0.703 \\

-w Euc & 0.835 & 0.910 & 0.846 & 0.695 \\ \hline

\end{tabular}}
\end{table}

\subsection{Sensitivity of Hyperparameter $\beta$ (RQ4)}
Hyperparameter $\beta$ is used to decide how close the $E_{str}$ and $E_{seq}$ should be, playing a crucial role in harmonizing the graph and sequence feature. Therefore, we analyze the influence of $\beta$ on detection performance in more detail. 

\textbf{Role of Hyperparameter $\beta$}. As shown in Figure 4, $\beta$ plays a critical role in ensuring robustness and generalizability across datasets. BREAK consistently achieves the best results across all datasets at $\beta$=0.1, where structural semantics are prioritized while retaining sufficient sequential cues. This setting enables effective semantic integration and demonstrates the necessity of a balanced denoising intensity. At $\beta$=0.6, the model shifts towards stronger structural denoising. While this benefits datasets with clearer semantic interrelations (e.g., PolitiFact), it disrupts the integration of sequential cues, which are critical for noisier datasets like Snopes and PolitiFact-S. This over-suppression weakens the model’s ability to infer nuanced relationships, leading to a performance drop. At $\beta$=0.7, the model reintroduces more sequential semantic information, striking a better balance between denoising and preserving complementary features. This adjustment restores meaningful connections and improves performance, particularly for datasets with mixed or noisy semantic patterns. The improvement at ($\beta$=0.7) demonstrates BREAK’s adaptability in rebalancing structural and sequential semantics.

\textbf{Dataset-Specific Sensitivities}. The sensitivity to $\beta$ varies across datasets, reflecting their differing semantic structures. GossipCop and PolitiFact exhibit stable performance across varying $\beta$ values due to their structured writing logic and clearer semantic relationships. BREAK performs consistently well on these datasets, regardless of minor variations in $\beta$. In contrast, Snopes and PolitiFact-S are noisier, reflecting that they are more sensitive to changes in $\beta$. The drop at $\beta$=0.6 reflects the over-suppression of sequential semantics, while the recovery at $\beta$=0.7 highlights the importance of preserving these cues to handle less structured content effectively.

\subsection{Case Study on Structure Denoising and Feature Diversification (RQ5)}
In this subsection, we present more analysis on structure denoising and the 
failed detection cases.

\textbf{Structure Denoising}. As shown in Figure \ref{fig6}, a majority of nodes disseminate their information to other nodes, exhibiting high out-degrees. From this observation, we can conclude that sentences with high in-degrees are likely to be topic sentences, as other sentences revolve around these focal points. Moreover, we observe that sentences with higher weights predominantly function as factual statements, resulting in higher in-degrees. Conversely, sentences with lower weights play a supporting role, leading to higher out-degrees. It's noteworthy that news articles tend to reiterate facts, such as ``effective immediately'' and ``change from 185\% to 200\%'', contributing to the high similarity among sentences.

\textbf{Failed Detection Cases}. To identify where BREAK struggles, we analyzed failed detection cases from the PolitiFact-S dataset where BREAK failed but LLaMA2-7B succeeded. The performance comparison results are listed in Table \ref{table3_}, and the analysis reveals: a) BREAK struggles with extremely short news content (e.g., a single sentence), where the lack of contextual richness makes it challenging to infer strong relationships between evidence and content. This limitation is particularly evident in datasets like PolitiFact-S, which contain a higher proportion of shorter articles; b) These cases align with BREAK’s design motivation, which is optimized for standard news articles comprising multiple sentences. Such articles provide richer structural and sequential semantics, which BREAK effectively captures. Addressing short-content scenarios, however, remains an area for future improvement.

\section{A Brief Analysis of the Innovation of Our Approach}
To help better understand, we further expounded on the innovation in our paper in a more concise way and elaborated on it from both theoretical and experimental aspects.

\textbf{Existing Problems}. Compared to other types of text, news articles often reiterate key facts, leading to intricate and dense semantic interrelations and high semantic similarity between sentences. These characteristics can interfere with distinguishing key information, acting as noise in the context of fake news detection. Specifically, Structural noise (semantic interrelations) can hinder traditional methods by obscuring key semantic interrelations within dense connections, forcing approaches like \cite{www2024msynfd} and \cite{22WWWrumor} to rely on small sliding windows, which limits their ability to capture global context. Feature noise, characterized by high similarity between sentences, can lead to vagueness in information during message passing, prompting methods like \cite{21Compare}, \cite{vaibhav2019}, and \cite{2023TCSS} to discard sentence semantics. These challenges result in traditional methods overlooking long-range semantics or sentence-level distinctions, leading to suboptimal performance.

\textbf{Theoretical Innovations}. a) We devise bi-level Framework for \textbf{Simultaneous} Denoising of Structure and Features: The innovation lies in the interdependent optimization of graph structure and features. Unlike traditional methods that treat structure and feature denoising independently, BREAK establishes a feedback loop where the improvement in one dimension (e.g., graph structure) enhances the other (e.g., graph features) iteratively. This results in a more robust representation that reduces the error propagation common in sequential processing. b) We address the challenge of integrating heterogeneous semantics (graph and sequence representations) with differing representation spaces for graph and sequence. Direct concatenation often yields suboptimal results due to the mismatch in space metrics. To overcome this, we introduce an affinity-based alignment mechanism, which creates a unified "golden representation", enabling effective integration. The novelty lies in using semantic affinity as a bridge to harmonize heterogeneous representations, which is validated experimentally (e.g., "-w/o seq" and "-w/o gra" variants show significant performance degradation). c) We devise a dynamic edge weight inference mechanism, which refines the graph structure by recalibrating edge weights based on semantic interrelations. The novelty lies in using node-level semantic refinement to dynamically guide structural adjustments, which improves noise resilience. The effectiveness is demonstrated by the significant gap between "-w/o inf" and the full model. d) We perform effectively semantic feature diversification. By enforcing KL-divergence, BREAK ensures that structural and sequential representations capture complementary information. This theoretical innovation enables BREAK to enhance feature robustness in noisy environments, as evidenced by "-w/o seq" results.

\textbf{Empirical Evidence Supporting Novelty}. The results illustrated in Figure \ref {fig3} highlight the necessity of each component of BREAK; any absence of these components would cause suboptimal results, thereby validating our theoretical contributions. 

In conclusion, the theoretical foundation of BREAK lies in its ability to tackle longstanding challenges in graph-based learning, such as noise resilience, heterogeneous semantic integration, and feature redundancy.

\end{document}